\documentclass[10pt,letterpaper]{article}
\usepackage{newtxtext}

\usepackage{amsthm, amssymb, amsmath, bm, bbm, mathtools, physics, dsfont}
\usepackage[scr=boondoxo]{mathalfa}
\usepackage[usenames,dvipsnames,svgnames,table]{xcolor}
\usepackage[pdftex, xetex]{graphicx}
\usepackage{float, colortbl, tabularx, longtable, multirow, subcaption, environ, wrapfig, textcomp, nicefrac,booktabs,colortbl,pdfpages,fontawesome,soul,comment, pgf, tikz,framed,tikz-cd,adjustbox,blindtext,setspace,multicol}
\usepackage{caption}
\usepackage[normalem]{ulem}
\usepackage[shortlabels,inline]{enumitem}
\usetikzlibrary{arrows,positioning,automata,shadows,fit,shapes,cd}
\usepackage{anyfontsize}

\usepackage[T1]{fontenc}
\usepackage{textcomp}
\usepackage[USenglish]{babel}
\usepackage[final]{microtype}

\usepackage[allcolors=black,colorlinks=true]{hyperref}
\usepackage{bookmark}
\usepackage{url}
\usepackage{tablefootnote}
\usepackage{makecell}
\usepackage[perpage]{footmisc}

\definecolor{myred}{RGB}{153,51,51}
\definecolor{mygray}{RGB}{102,102,102}
\definecolor{darkgreen}{RGB}{60,160,60}
\definecolor{lightblue}{RGB}{60,60,200}
\definecolor{darkred}{RGB}{70,10,10}
\definecolor{darkblue}{RGB}{1,1,255}
\definecolor{yellow}{RGB}{255,255,0}
\definecolor{purple}{RGB}{127,0,255}
\definecolor{dark}{RGB}{1,1,1}
\definecolor{lightgray}{RGB}{230,230,230}
\definecolor{ucla_gold}{RGB}{255,232,0}
\definecolor{ucla_blue}{RGB}{50,132,191}
\definecolor{lightyellow}{rgb}{1.0, 1.0, 0.88}

\usepackage[capitalize]{cleveref}
\crefformat{section}{Sec.\@ #2#1#3}
\crefmultiformat{section}{Sec.\@ #2#1#3}{ and~#2#1#3}{, #2#1#3}{, and~#2#1#3}
\crefformat{equation}{(#2#1#3)}
\renewcommand{\Cref}[1]{\cref{#1}}

\newtheoremstyle{mystyle}
  {}
  {}
  {\upshape}
  {}
  {\bfseries}
  {.}
  { }
  {\thmname{#1}\thmnumber{ #2}\thmnote{ (#3)}}

\theoremstyle{mystyle}

\theoremstyle{mystyle}

\makeatletter

\makeatletter

\definecolor{color_skyblue}{rgb}{0.01,0.39,0.75}

\usepackage{chngpage,tablefootnote,authblk,lineno}
\usepackage[title]{appendix}

\usepackage[small]{titlesec}

\usepackage[autocite=superscript, backend=bibtex, sortcites=true, sorting=none, bibstyle=nature, citestyle=numeric-comp, maxbibnames=15]{biblatex}
\renewbibmacro{in:}{}

\makeatletter
\newcommand{\printfnsymbol}[1]{%
  \textsuperscript{\@fnsymbol{#1}}%
}
\makeatother

\usepackage[color=gray!30,textsize=tiny,disable]{todonotes}

\date{}

\begin{document}
\listoftodos
\clearpage

\title{\Large
Bias in Machine Learning Models Can Be Significantly Mitigated by Careful Training: Evidence from Neuroimaging Studies
}
%
\author[1,2,3]{\normalsize Rongguang Wang}
\author[1,4]{Pratik Chaudhari\printfnsymbol{1}\printfnsymbol{2}}
\author[1,2,3,5]{Christos Davatzikos\footnote{Corresponding authors. Email: \href{mailto:pratikac@seas.upenn.edu}{pratikac@seas.upenn.edu}; \href{mailto:christos.davatzikos@pennmedicine.upenn.edu}{christos.davatzikos@pennmedicine.upenn.edu}. Address: 200 S 33rd Street, Department of Electrical and Systems Engineering, University of Pennsylvania, Philadelphia, PA 19104, USA; 3700 Hamilton Walk, 7th Floor, Center for Biomedical Image Computing and Analytics, University of Pennsylvania, Philadelphia, PA 19104, USA.}
\thanks{These authors contributed equally to this work.}
}
\author[ ]{\footnote{For the iSTAGING~\autocite{habes2021brain}{} and PHENOM~\autocite{chand2020two}{} consortia, and for the ADNI~\autocite{jack2008alzheimer}{}.}}
\affil[1]{\small Department of Electrical and Systems Engineering, University of Pennsylvania}
\affil[2]{Center for AI and Data Science for Integrated Diagnostics, University of Pennsylvania}
\affil[3]{Center for Biomedical Image Computing and Analytics, University of Pennsylvania}
\affil[4]{Department of Computer and Information Science, University of Pennsylvania}
\affil[5]{Department of Radiology, Perelman School of Medicine, University of Pennsylvania}
\maketitle

\begin{abstract}
Despite the great promise that machine learning has offered in many fields of medicine, it has also raised concerns about potential biases and poor generalization across genders, age distributions, races and ethnicities, hospitals, and data acquisition equipment and protocols. In the current study, and in the context of three brain diseases, we provide evidence which suggests that when properly trained, machine learning models can generalize well across diverse conditions and do not necessarily suffer from bias. Specifically, by using multi-study magnetic resonance imaging consortia for diagnosing Alzheimer’s disease, schizophrenia, and autism spectrum disorder, we find that well-trained models have a high area-under-the-curve (AUC) on subjects across different subgroups pertaining to attributes such as gender, age, racial groups, and different clinical studies and are unbiased under multiple fairness metrics such as demographic parity difference, equalized odds difference, equal opportunity difference etc. We find that models that incorporate multi-source data from demographic, clinical, genetic factors and cognitive scores are also unbiased. These models have better predictive AUC across subgroups than those trained only with imaging features but there are also situations when these additional features do not help.

\end{abstract}

\section{Introduction}

Machine learning models have shown great promise for precision diagnosis, treatment prediction, and a number of other clinical applications~\autocite{myszczynska2020applications}{}. There is increasing interest in building systems where such models can aid human experts in clinical settings. However, there are some key challenges to achieving this goal~\autocite{rajpurkar2022ai}{}. Clinical data is highly heterogeneous. For neurological disorders such as Alzheimer’s disease, it stems not only from diverse anatomies, overlapping clinical phenotypes, or genomic traits of different subjects, but also from operational, demographic and social aspects such as data acquisition protocols~\autocite{pomponio2020harmonization} and paucity of data for minorities. As a consequence, machine learning models often have poor reproducibility across different subgroups in the population.

We focus on one particular aspect of this issue, namely that models make biased predictions, i.e., they have different AUCs  for different genders, age and racial groups, and cohorts from different clinical studies. This has received wide attention~\autocite{larrazabal2020gender, gao2020deep, seyyed2021underdiagnosis, finlayson2021clinician, li2022cross, petersen2022feature, stanley2022fairness}{}. For example, a model trained on X-ray images consistently made inaccurate predictions on underrepresented genders when the training data was imbalanced~\autocite{larrazabal2020gender}{}. Similarly, for classification of chest X-ray pathologies, minority groups such as female African-Americans and patients with low socioeconomic status are prone to be incorrectly diagnosed as healthy~\autocite{seyyed2021underdiagnosis}{}. Such results have raised concerns on whether machine learning models can provide unbiased predictions, and whether they can be deployed in clinical settings eventually.

We provide results from relatively large and diverse datasets pertaining to three neurological disorders, namely Alzheimer’s disease (AD), schizophrenia (SZ), and autism spectrum disorder (ASD), which partly alleviate these concerns. We build models using magnetic resonance (MR) imaging features along with demographic, clinical, genetic factors, and cognitive scores, from large-scale consortia---iSTAGING~\autocite{habes2021brain} for AD, PHENOM~\autocite{chand2020two} for SZ, and ABIDE~\autocite{di2014autism} for ASD; see details in~\cref{s:app:data}. There are large imbalances in this dataset, e.g., 13\% and 37.4\% female subjects respectively in ABIDE and PHENOM, and 70.6\% European Americans, 8.8\% African Americans, and 1.5\% Asian Americans in iSTAGING. A large body of literature has studied confounders such as gender, age, race, image acquisition protocol for these data~\autocite{chyzhyk2022remove, benkarim2022population}{}. We show that, when trained with appropriate data pre-processing techniques and hyper-parameter tuning, machine learning models do not have biased predictions across different subgroups. In contrast, a baseline deep neural network provides accurate predictions on-average across the population but suffers from bias.

\begin{figure}[!p]
\centering
\includegraphics[width=\linewidth]{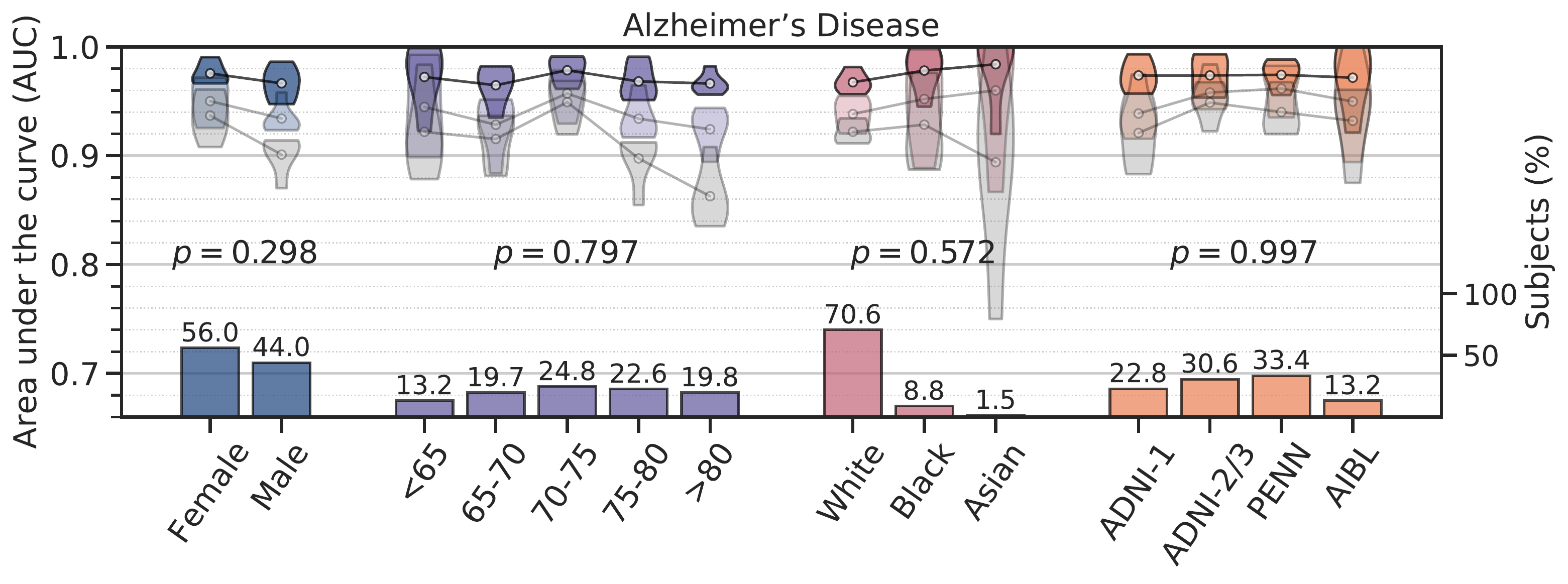}\\[-0.5em]
\includegraphics[width=\linewidth]{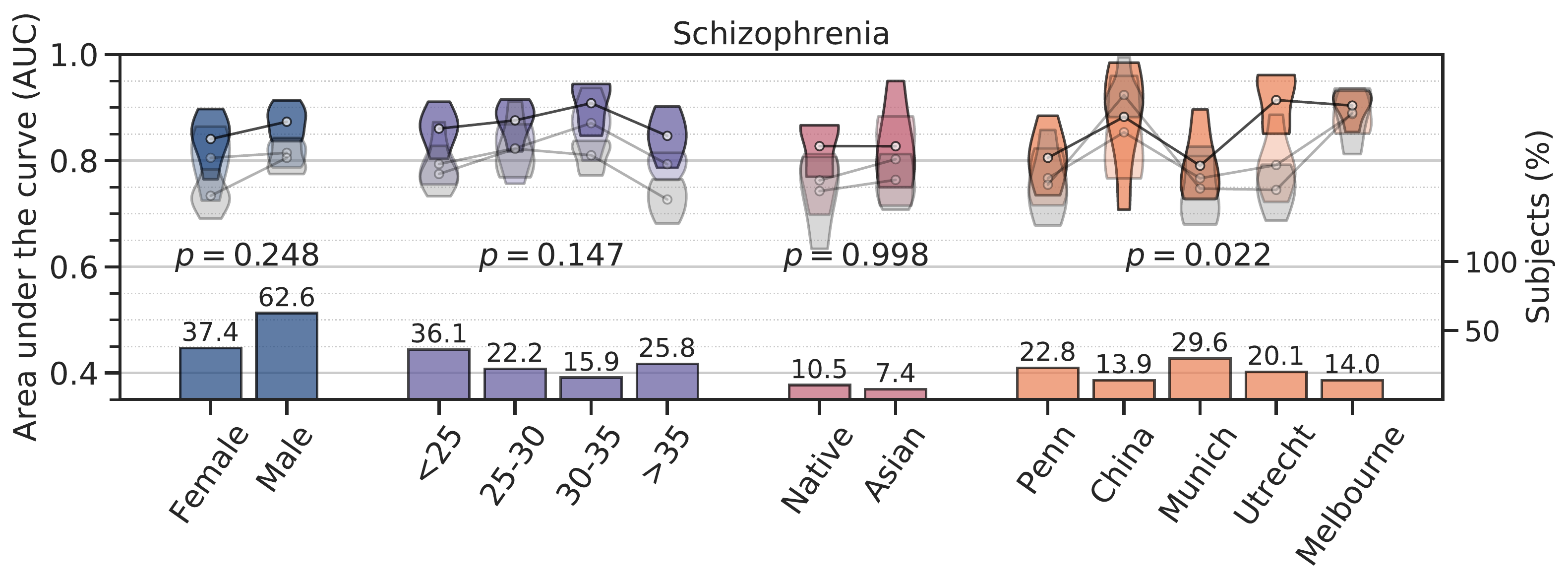}\\[-0.5em]
\includegraphics[width=\linewidth]{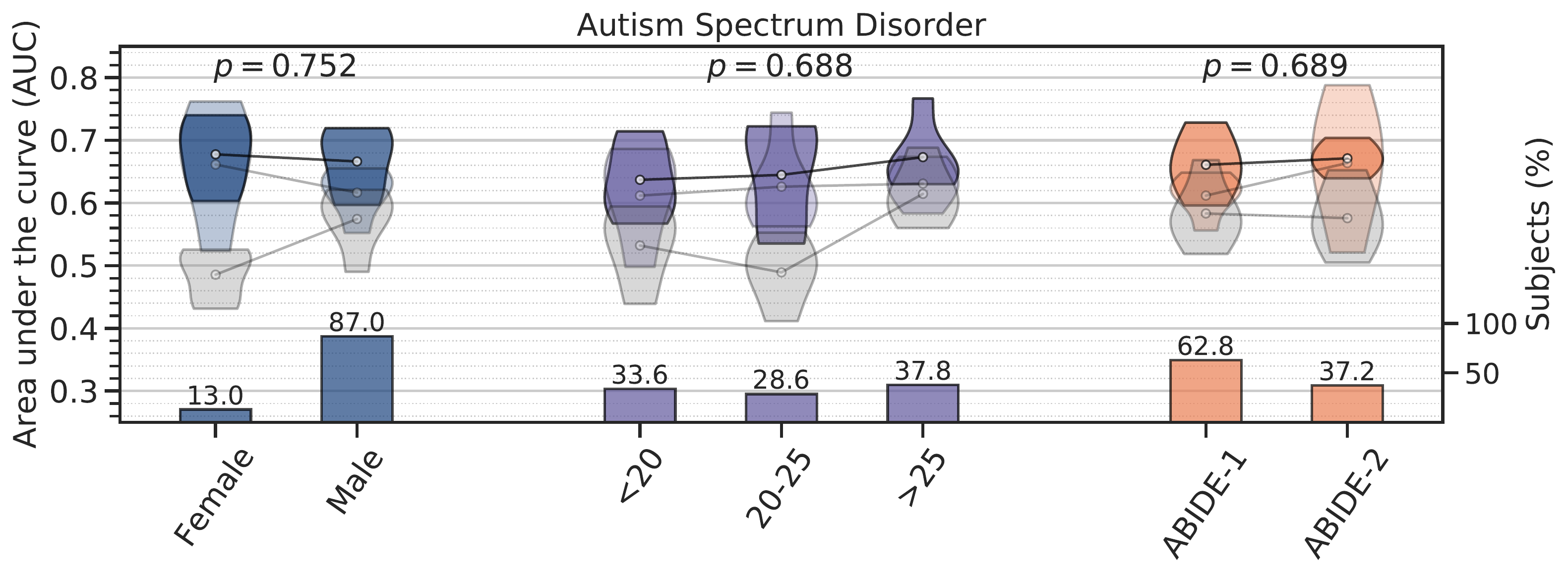}
\caption{
\textbf{Evaluating machine learning models on subjects from different gender, age, racial groups, and clinical studies.} For Alzheimer's disease, schizophrenia and autism spectrum disorder, using data from different studies (e.g., ADNI-1, ADNI-2/3, PENN and AIBL for Alzheimer's disease), we built an ensemble that uses data from multiple sources (MR imaging features, demographic, clinical variables, genetic factors, and cognitive scores). See details in~\cref{s:app:methods}. This ensemble was trained using pre-processing and hyper-parameter optimization discussed in~\cref{s:methods}. We also trained an ensemble using only imaging features. Bar plots denote the size of the subgroup/study (\%); in many cases, there is strong imbalance in the data. Violin plots denote the test AUC on five different held-out subsets of data. Solid colors indicate that models used all features while translucent colors indicate that models were trained only on imaging features. Translucent gray denotes the AUC of a baseline deep network (without appropriate preprocessing and hyper-parameter tuning). White dots denote the average AUC of each subgroup/study. For the ensemble trained on multi-source data, the $p$-values shown in the figure indicate that we cannot reject the null hypothesis that the AUC for different subgroups has the same mean (at significance level < 0.01). This is not the case for the baseline deep network.
}
\label{fig:all_plots}
\end{figure}

\section{Results}

\subsection{We demonstrate bias in baseline models in spite of their high accuracy}

The AUC of a deep network on held-out data (imaging features) is 0.924 ± 0.008 for AD, 0.806 ± 0.036 for SZ, and 0.572 ± 0.031 for ASD. These numbers are comparable to published results~\autocite{wen2020convolutional, rozycki2018multisite, katuwal2015predictive}{}. As Fig.~\ref{fig:all_plots} shows, for all three disorders, we find large discrepancies in the AUC on different subgroups for all attributes. For example, for AD the model has a higher AUC on females than males ($p$ and Cohen's $f$-value 2.10$\times 10^{-3}$, 1.13); for SZ the model predicts more accurately for males than females ($p$ and $f$-value 7.47$\times 10^{-3}$, 1.26). One may be inclined to hypothesize that this bias in AUC comes from one subgroup having a larger sample size than the other. This hypothesis does not hold when predictions are stratified by age. AD subjects older than 80 years, SZ subjects older than 35 years, and ASD subjects younger than 20 years, have a lower AUC ($p$-values are 1.93$\times 10^{-3}$, 6.31$\times 10^{-3}$ and 1.07$\times10^{-3}$, and $f$-values are 1.12, 1.06, and 1.02 respectively) even if these subgroups are not the ones with the smallest sample size. The baseline deep network is biased ($p$-value < 0.01) for all three disorders except in four cases: race and clinical studies in AD, race in SZ, and clinical studies in ASD.

\subsection{We demonstrate how this bias is significantly reduced when the models are carefully trained}

AUC of the ensemble (see Supplementary Information) on held-out data (imaging features) is 0.935 ± 0.014 for AD, 0.811 ± 0.026 for SZ and 0.631 ± 0.028 for ASD; all three are slightly better than that of the baseline deep network. As Fig.~\ref{fig:all_plots} shows visually, the AUC of the ensemble is more consistent. For all three neurological disorders, for all attributes, we cannot reject the hypothesis that AUCs for different subgroups have the same mean (at significance level < 0.01). In other words, the ensemble does not exhibit a bias, up to statistically indistinguishable levels. It is remarkable that this holds even with extreme imbalance in the data, e.g., 87\% males and 13\% females in ASD.

Using the same preprocessing and hyper-parameter tuning methodology as that of the ensemble, the bias of the baseline deep network can be improved. We obtained an AUC of 0.924 ± 0.013 on AD, 0.762 ± 0.023 on SZ and 0.565 ± 0.063 on ASD. These numbers, except for SZ, are about the same as that of the baseline deep network. The $p$-values (and Cohen's $f$-values in parentheses) for there being bias across subgroups now are 0.086 (sex, 0.69), 0.379 (age, 0.47), 0.976 (race, 0.06) and 0.578 (study, 0.36) for AD; 0.710 (sex, 0.14), 0.173 (age, 0.59), 0.776 (race, 0.10) and 4.77$\times 10^{-4}$ (study, 1.27) for SZ; 0.321 (sex\, 0.37), 0.254 (age, 0.51) and 0.818 (study, 0.08) for ASD. The deep network trained with better data preprocessing achieves a similar average AUC as that of the baseline network but its predictions are not biased (significance level < 0.01); except across clinical studies in SZ, which could be due to differences in clinical characteristics across patient cohorts.

\begin{figure}[!t]
\centering
\includegraphics[width=0.75\linewidth]{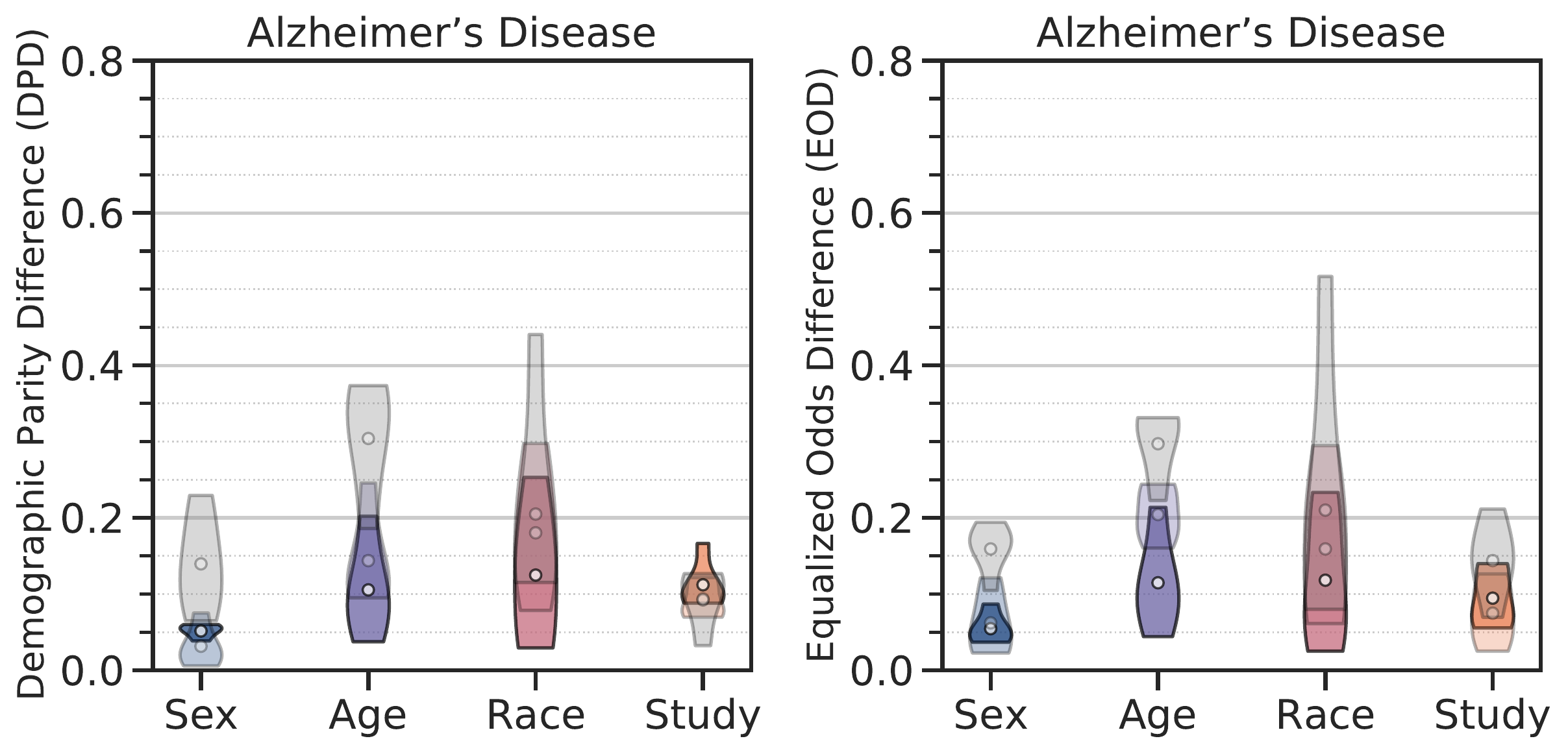}\\
\includegraphics[width=0.75\linewidth]{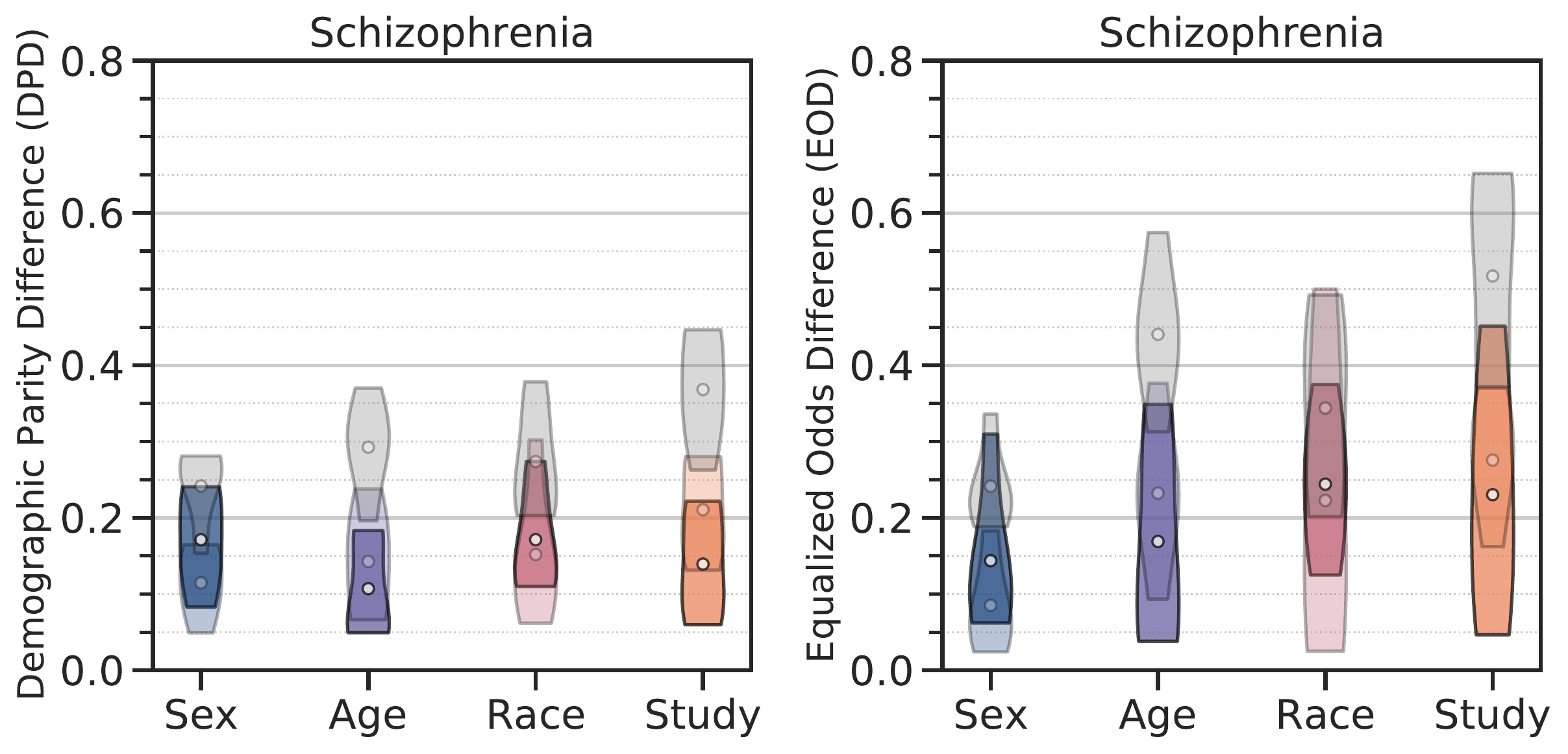}\\
\includegraphics[width=0.75\linewidth]{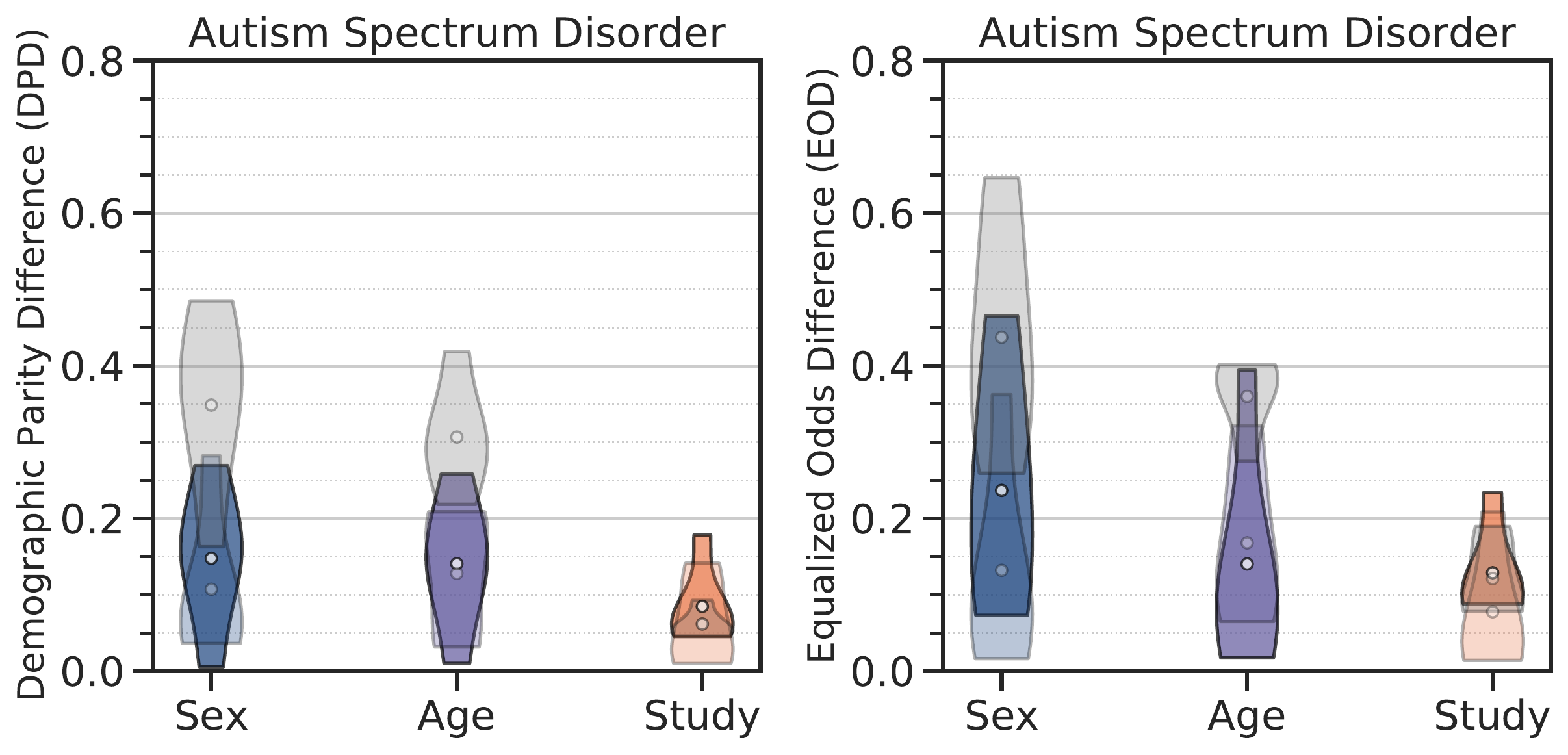}
\caption{
\textbf{Fairness assessment with respect to sensitive attributes.} Violin plots denote the test demographic parity differences (DPD) and equalized odds difference (EOD)~\autocite{agarwal2018reductions} on five different held-out subsets of data. White dots denote the average metric of each subgroup/study. A model is perfectly fair if the fairness disparity (DPD or EOD) is zero. Ensemble models (imaging features or multiple source) have lower disparities for all sensitive attributes in both metrics (differences in the two are not statistically significant) as compared to the baseline deep network (all $p$-values < 0.01). It has been noticed that different fairness metrics are often incompatible. We therefore also assessed fairness using three other metrics: equal opportunity difference (EO), predictive parity difference (PPD) and generalized entropy index (GEI). In all cases, ensemble models show lower disparities compared to baseline models, except the GEI metric for autism spectrum disorder (see~\cref{s:app:fair}).
}
\label{fig:fair_plots}
\end{figure}

\subsection{We also demonstrate how this bias is significantly reduced when models are trained on multi-source data}

We next trained the ensemble using demographic features, clinical variables, genetic factors and cognitive scores in addition to MR imaging features. The AUC of this multi-source ensemble on held-out data is 0.968 ± 0.009 for AD, 0.866 ± 0.024 for SZ and 0.663 ± 0.034 for ASD. All three values are better than corresponding ones for the ensemble trained only on structural MR imaging features ($p$-value < 3.61$\times 10^{-3}$ for all three). Therefore, using multi-source data improves the average AUC of models for these three disorders. This ensemble also makes unbiased predictions across different subgroups; the $p$-values (and $f$-values in parentheses) are 0.298 (sex, 0.39), 0.797 (age, 0.29), 0.572 (race, 0.31), and 0.997 (study, 0.05) for AD; 0.248 (sex, 0.44), 0.147 (age, 0.62), 0.998 (race, 0.00), and 0.022 (study, 0.86) for SZ; 0.752 (sex, 0.12), 0.688 (age, 0.25), and 0.689 (study, 0.15) for ASD.

A two-way ANOVA as to whether the improved average AUC translates to improved AUC for subgroups pertaining to each attribute has $p$-values 9.26$\times 10^{-3}$ (sex), 1.90$\times 10^{-2}$ (age), 0.465 (race), and 0.636 (study) for AD; 0.250 (sex), 3.31$\times 10^{-2}$ (age), 0.713 (race), and 7.86$\times 10^{-3}$ (study) for SZ; 4.24$\times 10^{-2}$ (sex), 0.237 (age), and 0.681 (study) for ASD. At a significance level of 0.01, we find that using multi-source data improves AUC of the ensemble as compared to using features from only imaging features for subgroups pertaining to sex in AD and clinical studies in SZ, but this does not hold for other cases.

\section{Discussion}

As machine learning is being applied to problems in the clinical sciences, there is an increasing amount of discussion on bias in particular, and ethical issues in general~\autocite{finlayson2021clinician}{}. There is also a large amount of recent work on identifying biases~\autocite{larrazabal2020gender, seyyed2021underdiagnosis, li2022cross}{}, and developing techniques to mitigate them~\autocite{gao2020deep, wang2022embracing}{}. We showed that when models are trained using appropriate data preprocessing and hyper-parameter optimization techniques, their predictions need not be biased for neuroimaging data. Baseline models, e.g., deep networks, that do not use this preprocessing and hyper-parameter optimization predict accurately on average but can be biased. We have also checked that training with sufficient data preprocessing but without adequate hyper-parameter tuning produces unbiased models but with lower AUCs (see~\cref{s:app:recomm}). Our results do not diminish the value of the existing work on bias. Instead, they provide evidence that we might be able to develop unbiased machine learning-based diagnostic models, and deploy them in practice in the future. However, in some situations, bias cannot be removed by preprocessing and model selection because the source of the bias may be somewhere else, e.g., different rates of misdiagnosis~\autocite{obermeyer2019dissecting}{}.

The machine learning literature has well-established safeguards against poor generalization. Disregarding these procedures can lead to biased predictions. Our results indicate that a rigorous preprocessing, training and evaluation methodology \emph{can} give models that do not suffer from biased predictions in neuroimaging applications. This has also been noticed recently~\autocite{petersen2022feature}{}, but we have demonstrated this phenomenon more exhaustively and across three different neurological disorders. Our work can therefore provide a benchmark for automated neuroimaging-based diagnostic systems.

Balanced datasets are desirable for building unbiased models~\autocite{chawla2002smote}{}. But it is extremely difficult to obtain balanced data. There are long-standing problems in recruiting volunteers across gender, age, and race. For example, female, minority ethnicity groups, and older subjects are less likely to participate in clinical trials~\autocite{chastain2020racial}{}. Even for balanced datasets, trained models may still be biased due to unobserved confounders, e.g., severity of the disease or genetic factors. Recent studies have therefore argued for training models on extensive multi-source neuroimaging datasets~\autocite{he2022meta}{}. Our results agree with such motivations---large-scale cohorts of multi-source data can enable training robust and unbiased models and also enable thorough evaluation.

\section{Materials and Methods}
\label{s:methods}

We use 3D magnetic resonance images along with demographic (gender, age, race, education level, marital status, employment status, handedness, smoker), clinical (diabetes, hypertension, hyperlipidemia, systolic/diastolic blood pressure and body mass index), and genetic factors (apolipoprotein E (APOE) alleles 2, 3 and 4) and cognitive scores from three large consortia---iSTAGING for AD, PHENOM for SZ, and ABIDE for ASD; see details in~\cref{s:app:data}. We use a standard processing pipeline (see~\cref{s:app:preprocess}) to compute imaging features from T1-weighted MR images. All accuracies are calculated using 5 independent held-out test sets.

\textbf{Feature pre-processing pipeline.}
Some features, predominantly clinical and genetic factors and cognitive scores, are sparsely populated; see details in~\cref{s:app:data}. Continuous-valued features are normalized to have zero mean and unit variance after median imputation; quantile normalization is used for features with skewed distributions. For categorical features, we introduce an ``unknown'' category for missing values and for each feature with missing values we introduce an additional Boolean feature which indicates whether the value was missing. No harmonization tools~\autocite{pomponio2020harmonization} are used.

We use AutoGluon~\autocite{erickson2020autogluon} for training all models, for building ensembles using bagging, boosting and stacking, and for performing hyper-parameter search. The baseline deep network has three fully-connected layers and is trained using data that is normalized to zero mean and unit variance after dropping missing values, without feature pre-processing.

\section*{Acknowledgments}
This work was supported by the National Institute on Aging (RF1AG054409 and U01AG068057), the National Institute of Mental Health (R01MH112070), the National Institutes of Health (75N95019C00022), the National Science Foundation (2145164) and cloud computing credits from Amazon Web Services.

\printbibliography[title=Bibliography]

\clearpage

\renewcommand\appendixname{Supplementary}
\begin{appendices}

\section{Data}
\label{s:app:data}

We use 3D magnetic resonance (T1-weighted) images along with demographic (gender, age, race, education level, marital status, employment status, handedness, smoker), clinical (diabetes, hypertension, hyperlipidemia, systolic/diastolic blood pressure and body mass index), genetic factors (apolipoprotein E (APOE) alleles 2, 3 and 4), and cognitive scores (mini-mental state exam (MMSE), full-scale intelligence quotient (FIQ), verbal intelligence quotient (VIQ), and performance intelligence quotient (PIQ)) from three large consortia---iSTAGING~\autocite{habes2021brain} for AD, PHENOM~\autocite{satterthwaite2010association, zhang2015heterogeneity, chand2020two} for SZ, and ABIDE~\autocite{di2014autism} for ASD. The subset of iSTAGING data used here consists of multiple clinical studies: AD Neuroimaging Initiative (ADNI)~\autocite{jack2008alzheimer}{}, Penn Memory Center cohort (PENN), and Australian Imaging, Biomarkers and Lifestyle (AIBL)~\autocite{ellis2010addressing}{}. In PHENOM scans are acquired from five different sites namely Penn (United States), China, Munich, Utrecht, and Melbourne. All tasks in this study are a binary classification problem with two labels (healthy controls and patients). We only use the baseline (first time point) scans from each cohort; all follow-up sessions are excluded. This way there is no data leakage for the same participant between training and test sets. For AD, we select stable cognitive normal (CN) and AD patients based on each participant's longitudinal diagnosis status. We only include subjects who were diagnosed as CN or AD at the baseline and stayed stable during the follow-up sessions.

\textbf{iSTAGING.} The consortium consists of four datasets including ADNI-1 (22.81\%), ADNI-2 (30.58\%), PENN (33.40\%), and AIBL (13.22\%). There are in total 781 controls and 815 patients where 173 (controls)/191 (patients) in ADNI-1, 261/227 in ADNI-2/3, 228/305 in PENN, and 119/92 in AIBL. 56.02\% of the participants are female and 43.98\% of them are male; 10.90\% (female)/11.90\% (male) in ADNI-1, 15.66\%/14.91\% in ADNI-2/3, 21.37\%/12.03\% in PENN, and 8.08\%/5.14\% in AIBL. The participants are distributed in age ranges 0--65 (13.16\%), 65--70 (19.67\%), 70--75 (24.81\%), 75--80 (22.56\%), and over 80 (19.80\%) years old; 1.50\% (0--65)/1.94\% (65--70)/7.27\% (70--75)/6.52\% (75--80)/5.58\% (> 80) in ADNI-1, 3.51\%/8.15\%/6.95\%/6.89\%/5.08\% in ADNI-2/3, 6.33\%/7.02\%/6.83\%/6.58\%/6.64\% in PENN, and 1.82\%/2.57\%/3.76\%/2.57\%/2.51\% in AIBL. 70.61\% of the participants are White, 8.77\% of them are Black, and 1.50\% of them are Asian; 21.18\% (White)/1.19\% (Black)/0.31\% (Asian) in ADNI-1, 16.98\%/0.88\%/0.56\% in ADNI-2/3, 25.13\%/6.70\%/0.63\% in PENN, and 7.33\% (White) in AIBL.

\textbf{PHENOM.} The consortium consists of five datasets including Penn (22.28\%), China (13.94\%), Munich (29.64\%), Utrecht (20.12\%), and Melbourne (14.03\%). There are in total 563 controls and 456 patients where 131 (controls)/96 (patients) in Penn, 76/66 in China, 157/145 in Munich, 115/90 in Utrecht, and 84/59 in Melbourne. 37.39\% of the participants are female and 62.61\% of them are male; 11.87\% (female)/10.40\% (male) in Penn, 6.77\%/7.16\% in China, 7.75\%/21.88\% in Munich, 6.97\%/13.15\% in Utrecht, and 4.02\%/10.01\% in Melbourne. The participants are distributed in age ranges 0--25 (36.11\%), 25--30 (22.18\%), 30--35 (15.90\%), and over 35 (25.81\%) years old; 5.79\% (0--25)/6.28\% (25--30)/3.53\% (30--35)/6.67\% (> 35) in Penn, 4.91\%/2.36\%/2.16\%/4.51\% in China, 9.42\%/7.26\%/5.99\%/6.97\% in Munich, 10.11\%/4.12\%/2.85\%/3.04\% in Utrecht, and 5.89\%/2.16\%/1.37\%/4.61\% in Melbourne. 10.50\% of the participants are Native and 7.36\% of them are Asian; 10.50\% (Native)/7.36\% (Asian) in Penn.

\textbf{ABIDE.} The consortium consists of two phases including ABIDE-1 (62.78\%) and ABIDE-2 (37.22\%). There are in total 362 controls and 307 patients where 224 (controls)/196 (patients) in ABIDE-1 and 138/111 in ABIDE-2. 13.00\% of the participants are female and 87.00\% of them are male; 6.73\% (female)/56.05\% (male) in ABIDE-1 and 6.28\%/30.94\% in ABIDE-2. The participants are distributed in age ranges 0--20 (33.63\%), 20--25 (28.55\%), and over 25 (37.82\%) years old; 21.97\% (0--20)/17.49\% (20--25)/23.32\% (> 25) in ABIDE-1 and 11.66\%/11.06\%/14.50\% in ABIDE-2.

\textbf{Available variables.} In the iSTAGING consortium, we have MR imaging (region-of-interest volumes and white matter lesion volume), demographics (gender, age, race and smoking status), clinical (diabetes, hypertension, hyperlipidemia, blood pressure (systolic/diastolic) and body mass index), genetic factor (Apolipoprotein E alleles 2, 3 and 4), and cognitive score (mini-mental state exam) variables. In the PHENOM consortium, we have MR imaging (region-of-interest volumes) and demographics (gender, age, race, education level, marital status, employment status and handedness) variables.
In the ABIDE consortium, we have MR imaging (region-of-interest volumes), demographics (gender, age and handedness) variables, and cognitive score (full-scale intelligence quotient, verbal intelligence quotient and performance intelligence quotient) variables.

\section{Methodology for creating features from structural measures}
\label{s:app:preprocess}

We compute features from T1-weighted MR images using a standard pipeline. Scans are bias-field corrected~\autocite{tustison2010n4itk}{}, skull-stripped with a multi-atlas algorithm~\autocite{doshi2013multi}{}, and then a multi-atlas label fusion segmentation method~\autocite{doshi2016muse}{} is used to obtain anatomical region-of-interest (ROI) masks for 119 gray matter ROIs, 20 white matter ROIs and 6 ventricle ROIs of the brain (total 145). We further segment white matter hyperintensities (WMH) using a deep learning-based algorithm~\autocite{doshi2019deepmrseg} on fluid-attenuated (FLAIR) and T1-weighted images. White matter lesion (WML) volumes are obtained by summing up the WMH mask voxels.

\textbf{Feature pre-processing pipeline.}
Our data contains structural features such as ROI and WML volumes in addition to demographic, clinical and genetic factors, and cognitive scores. Some of these features (predominantly the last three) are sparsely populated. 
For continuous-valued features, we first impute missing values with the median of each variable and normalize the feature to have zero-mean and unit-variance. We apply quantile normalization to skewed distributions. For discrete-valued features, we introduce a ``unknown'' category for missing values. Corresponding to each feature with missing values, we introduce an additional Boolean feature which indicates whether the value was missing. This way we preserve the evidence of absence (rather than the absence of evidence)~\autocite{erickson2020autogluon}{}. We did not use any harmonization tools~\autocite{pomponio2020harmonization, wang2021harmonization}{}.

\section{Evaluation methodology}
We report area under the receiver
operating characteristic (AUC) curve on held-out test sets as follows. We split data into 5 equal-sized folds (stratified by labels), use four for training and validation (80\%) and the fifth for testing (20\%). All hyper-parameter tuning is performed using a further 5-fold cross-validation within the 80\% data. This way, the 20\% data is a completely independent test set which is used only for reporting the final AUC. We report mean and standard deviation of the AUC over 5 independent test sets (one for each outer fold). This is a computationally expensive, but rigorous, evaluation methodology. The three neurological disorders consist of data from multiple clinical studies; we create the training, validation and test sets for each study independently and then concatenate them.

\section{Hyper-parameter tuning methodology}
\label{s:app:methods}

We compare results from an optimized model, in which hyper-parameter optimization and ensemble learning were performed, with a basic network. For the former, we use a framework called AutoGluon~\autocite{erickson2020autogluon} which gives an easy way to train a large number of different types of models (deep networks, $k$-nearest neighbor classifiers, random forests, CatBoost~\autocite{prokhorenkova2018catboost}{}, and LightGBM~\autocite{ke2017lightgbm}) and perform hyper-parameter search. For deep network models we create an input layer that concatenates the embedding of continuous-valued and categorical features; the other models can natively handle both these types of features. Using AutoGluon we can also build ensembles of these models via bagging, boosting and stacking. For each neurological disorder, for each of the 5 outer folds, we train the above different types of models using different hyper-parameters in parallel across multiple CPUs and 4 GPUs for 1 hour and build an ensemble that obtains the best classification log-likelihood on the validation data.

The baseline deep network has three fully-connected layers and is also built within the same software framework. This network is trained using data that is normalized to have zero mean and unit standard deviation after dropping missing values. It does not use the pre-processing pipeline described above.

\section{Evaluation on additional fairness metrics}
\label{s:app:fair}

It has been noticed that different fairness metrics are often incompatible. We therefore also assessed fairness using three other metrics. Equal opportunity difference (EO)~\autocite{hardt2016equality}{}, predictive parity difference (PPD)~\autocite{garg2020fairness}{} and generalized entropy index (GEI)~\autocite{speicher2018unified} are three other well-known group fairness measures. A predictor satisfies ``equal opportunity'' if the true positive rate (TPR) is the same across sensitive groups. ``Predictive parity'' is satisfied when the positive predictive value (PPV) is the same for sensitive groups, where PPV is defined as the probability that individuals predicted to belong to the positive class actually belong to the positive class. ``Generalized entropy index'' is an unified individual and group fairness measure and it also explains how individual fairness and group fairness are related. A value of zero represents perfect equality and higher values denote increasing levels of inequality. We assess fairness of our learned learning models using EO, PPD and GEI with respect to sensitive groups including sex, age, race, and study for all neurological disorders. Our original findings remain unchanged: appropriately designed machine learning models show lower disparities compared to baseline machine learning models under these metrics (except GEI metric for autism spectrum disorder case).

For all three neurological disorders, we find that the ensemble model has small fairness disparities in terms of EO, PPD and GEI on all sensitive groups. For example, for AD, the EO, PPD and GEI of the baseline deep network on held-out data are 0.159 ± 0.029, 0.159 ± 0.028 and 0.073 ± 0.005 across gender respectively, which are larger than those of an ensemble model, 0.043 ± 0.048, 0.029 ± 0.028 and 0.058 ± 0.005 respectively ($p$-values 3.23$\times 10^{-3}$, 1.68$\times 10^{-4}$ and 2.62$\times 10^{-3}$). For SZ, the ensemble shows lower EO, PPD and GEI disparities (0.039 ± 0.042, 0.112 ± 0.060 and 0.102 ± 0.018) compared to the neural net (0.222 ± 0.022, 0.272 ± 0.098 and 0.142 ± 0.024) across age groups with $p$-values 9.02$\times 10^{-3}$, 2.37$\times 10^{-3}$ and 3.04$\times 10^{-3}$ respectively. Under EO and PPD metrics, compared to the baseline deep network, the ensemble model has better fairness ($p$-value < 0.01) for all three disorders except in four cases: race and clinical studies in AD, race in SZ, and clinical studies in ASD. Under GEI metric, the ensemble model has better fairness ($p$-value < 0.01) across all sensitive attributes for AD and SZ. There's no statistically significant difference in GEI between neural net and ensemble for ASD. We also found that the differences between between ensembles trained only on structural measures and ones trained on multi-source data in terms of these fairness metrics are not statistically significant.

\section{Some practical recommendations for training unbiased machine learning models}
\label{s:app:recomm}

We conducted the following experiment for Alzheimer’s disease diagnosis using a deep neural net to understand the influence of data pre-processing and hyper-parameter tuning techniques on bias. We performed experiment under the following three settings.
\begin{enumerate}
\item[(a)] Training without sufficient data pre-processing but with adequate hyper-parameter tuning. This is the case we have shown in the first experiment (baseline deep network). We find that both gender and age sub-groups show biased predictions in this case.
\item[(b)] Training with sufficient data pre-processing but without adequate hyper-parameter tuning. We run hyper-parameter tuning for 10$\times$ less computer time compared to case (a). We observe that prediction disparity does not appear, and this holds for all sub-groups. However, this inadequate hyper-parameter tuning leads to relatively poor prediction performance (0.896 in AUC) compared to case (a) (0.924 in AUC).
\item[(c)] Training with both adequate data pre-processing and hyper-parameter tuning. This is the setting that we have used for the experiments with the ensemble. As we discussed in the main text, machine learning models do not exhibit biased predictions in this case. This setting also leads to improved AUC, which matches that of case (a).
\end{enumerate}
These ablation experiments suggest that adequate data pre-processing leads to unbiased models which obtain a high AUC. On the other hand, adequate hyper-parameter tuning ensures an accurate model, but it may not provide unbiased predictions. 
One may therefore ascribe importance to the various factors:data pre-processing, hyper-parameter tuning, ensembling, and multi-source data. Our suggestions for building unbiased and accurate predictive models are as follows.
\begin{itemize}
\item Use adequate data pre-processing and hyper-parameter tuning techniques;
\item Use ensemble models to obtain robust predictions; there are a number of effective techniques to mitigate to increase in computational complexity of ensembles~\autocite{chen2018tvm}{};
\item Leverage multi-source data if they are available~\autocite{acosta2022multimodal}{}.
\end{itemize}

\end{appendices}

\end{document}